%% file: acl.tex
\pdfoutput=1

\documentclass[11pt]{article}

\usepackage{acl}

\usepackage{times}
\usepackage{latexsym}

\usepackage[T1]{fontenc}

\usepackage[utf8]{inputenc}

\usepackage{microtype}

\usepackage{multirow}
\usepackage{pifont}
\usepackage{amsmath, amsfonts}
\usepackage{algorithm}
\usepackage{algorithmicx}
\usepackage{algpseudocode}
\usepackage{subfigure}

\usepackage{amsfonts}
\usepackage{amsmath}
\usepackage{graphicx}
\usepackage{multirow}
\usepackage{caption}
\usepackage{subfigure}
\usepackage{enumitem}
\usepackage{float}
\usepackage{makecell}
\usepackage{booktabs}
\usepackage{tabu}

\usepackage[T1]{fontenc}
\usepackage[utf8]{inputenc}
\usepackage{longtable}
\usepackage{booktabs} 
\usepackage{multirow}
\usepackage{amsmath}
\usepackage{booktabs}
\usepackage{amsfonts}
\usepackage{subfigure}
\usepackage{makecell}
\newcommand{\ours}{LightNER}

\newcommand{\ie}[0]{\emph{i.e., }}

\newcommand{\xv}{\boldsymbol{x}}
\newcommand{\yv}{\boldsymbol{y}}

\newcommand{\thetav}{\boldsymbol{\theta}}

\newcommand{\RN}[1]{%
	\textup{\lowercase\expandafter{\it \romannumeral#1}}%
}

\newcommand{\tableindent}{~~}

\title{LightNER: A Lightweight Tuning Paradigm for Low-resource NER via Pluggable Prompting}

\author{
	Xiang Chen$^{1,2}$\footnotemark[1] , Lei Li$^{1,2}$\thanks{$\quad$ Equal Contribution.} ,
	Shumin Deng$^{1,2}$, Chuanqi Tan$^{3}$, Changliang Xu\textsuperscript{\rm 4}\\
	\textbf{
	Fei Huang$^{3}$, Luo Si$^{3}$, Huajun Chen$^{1,2}$,
	Ningyu Zhang$^{1,2,}$\thanks{$\quad$ Corresponding Author.} 
	} \vspace{1.0mm} \\
	$^1$Zhejiang University \& AZFT Joint Lab for Knowledge Engine, China \\
	$^2$Hangzhou Innovation Center, Zhejiang University,
	$^3$Alibaba Group\\ $^4$State Key Laboratory of Media Convergence Production Technology and Systems \\
	\fontsize{11}{10}\selectfont 
	\{xiang\_chen, leili21, 231sm, huajunsir, zhangningyu\}@zju.edu.cn,  \\
	\fontsize{11}{10}\selectfont \{chuanqi.tcq, f.huang, luo.si\}@alibaba-inc.com, xu@shuwen.com \\
}

\begin{document}
\maketitle

\begin{abstract}
Most NER methods rely on extensive labeled data for model training, which struggles in the low-resource scenarios with limited training data. Existing dominant approaches usually suffer from the challenge that the target domain has different label sets compared with a resource-rich source domain, which can be concluded as class transfer and domain transfer. In this paper, we propose a lightweight tuning paradigm for low-resource NER via pluggable prompting (LightNER). Specifically,  we construct the unified learnable verbalizer of entity categories to generate the entity span sequence and entity categories without any label-specific classifiers, thus addressing the class transfer issue. We further propose a pluggable guidance module by incorporating learnable parameters into the self-attention layer as guidance, which can re-modulate the attention and adapt pre-trained weights. Note that we only tune those inserted module with the whole parameter of the pre-trained language model fixed, thus, making our approach lightweight and flexible for low-resource scenarios and can better transfer knowledge across domains. Experimental results show that LightNER can obtain comparable performance in the standard supervised setting and outperform strong baselines in low-resource settings\footnote{Code is in \url{https://github.com/zjunlp/DeepKE/tree/main/example/ner/few-shot}.}.

\end{abstract}

\input{intro}

\input{related}

\input{method}

\input{experiments}

\section{Conclusion and Future Work}
In this paper, we propose a lightweight tuning paradigm for low-resource NER via pluggable prompting (LightNER), which can
accomplish the class transfer and domain transfer for low-resource NER without modifying the PLM's parameters and architecture.
Note that we only tune pluggable guidance module with the whole parameter of the PLMs fixed, thus, making our approach lightweight and flexible for low-resource scenarios and can better transfer knowledge across domains and tasks.
Experimental results reveal that LightNER can obtain competitive results in the rich-resource setting and outperform baselines in the low-resource setting. 
In the future, we plan to leverage knowledge graphs to enhance the pluggable guidance module for better knowledge transfer performance.

\section*{Acknowledgments}

We  want to express gratitude to the anonymous reviewers for their kind comments. 
This work was supported by National Natural Science Foundation of China (No.62206246, 91846204 and U19B2027), Zhejiang Provincial Natural Science Foundation of China (No. LGG22F030011), Ningbo Natural Science Foundation (2021J190), and Yongjiang Talent Introduction Programme (2021A-156-G).

\section*{Ethics/Broader Impact Statement}
The ability to extract useful and important information from text, without the need for human input or control, has a wide range of practical and industrial applications.
Named Entity Recognition is an important information extraction task that can benefit many NLP applications, e.g., information retrieval, dialog generation, and question answering.
However, traditional fine-tuning-based approaches achieve poor performance in situations in which there are sparse data available. 
Our simple-yet-effective approach makes it possible to obtain better performance in these low-resource settings and we detail its advantages as:

(i) The few-shot setting is realistic (the number of labeled instance per class $K$ can be any variable);

(ii) No necessity of prompt engineering;

(iii) Extensible to any pre-trained LMs (e.g., BERT or GPT-2).

Our motivation is to develop an applicable, generalizable lightweight tuning paradigm
for the NLP community and our work is but a small step towards this direction.  

\bibliography{custom}
\bibliographystyle{acl_natbib}


\appendix
\input{appendix}

\end{document}

%% file: intro.tex
\section{Introduction}
Named Entity Recognition (NER) has been a fundamental task of research within the Natural Language Processing (NLP) community. 
Mostly, the NER task is formulated as a sequence classification task, aiming to assign the labels to each entity in the input sequence. 
And those entity labels are all based on pre-defined categories, such as location, organization, person. 
The current mature methodologies for handling NER is using Pre-trained language models (PLMs) ~\cite{DBLP:conf/naacl/DevlinCLT19} equipped with several NER paradigms to perform extensive training process on large corpus, such as label-specific classifier paradigm (LC)~\cite{dilated-convolutions,lan}, machine reading comprehension paradigm (MRC) ~\cite{DBLP:conf/acl/YuBP20} and unified generative paradigm (BartNER~\cite{unifiedNER}).
Unfortunately the resulting models are highly associated with seen categories, which often explicitly memorizing entity values~\cite{DBLP:journals/coling/AgarwalYWN21}, partially because the output layers require a consistent label set between training and testing. 
Note that these models require to  build a new model from scratch to adapt to a target domain with new entity categories, thus, achieving unsatisfactory performance when the target labeled data is limited.

\begin{figure}[t!]
    \centering
    \includegraphics[width=0.48\textwidth]{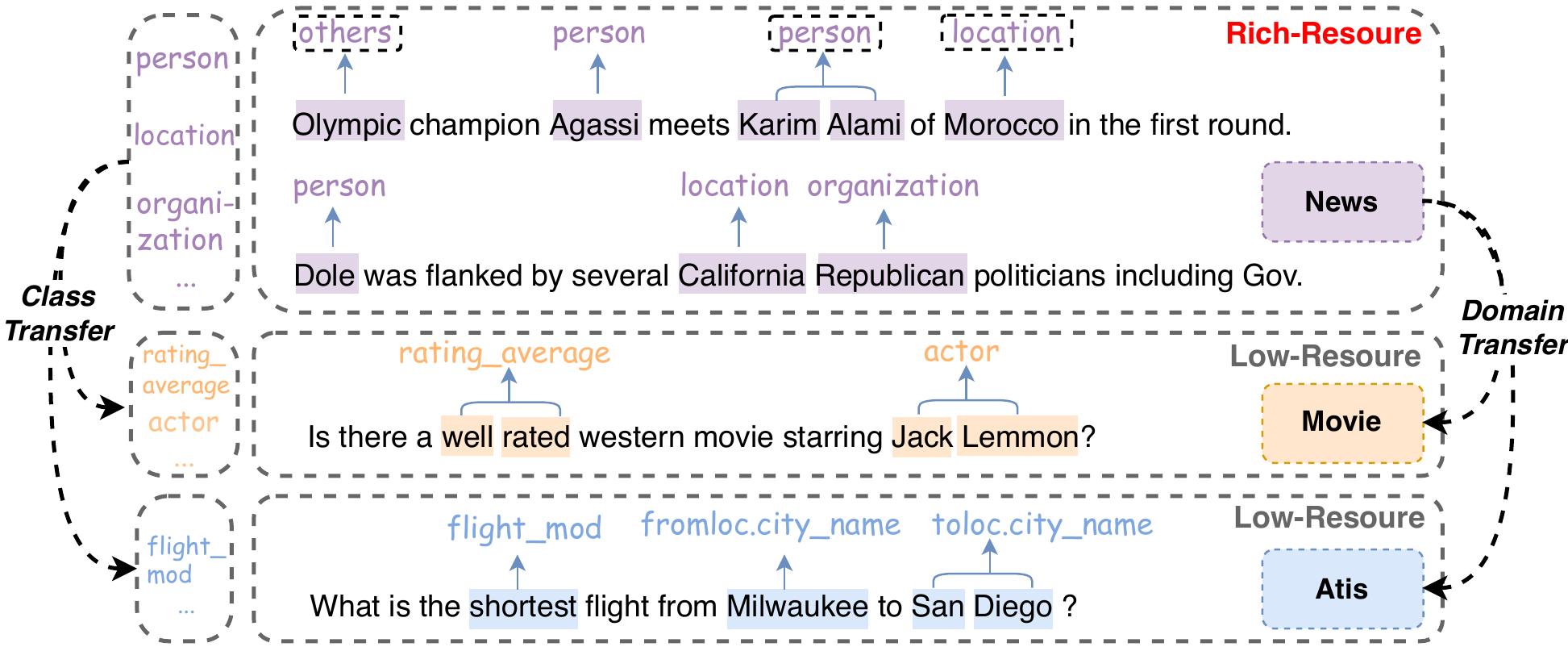}
    \caption{Examples of NER involved in \textbf{Class Transfer} and \textbf{Domain Transfer} in low-resource setting.}
    \label{fig:example}
\end{figure}

Unfortunately, this problem is prevalent in real-world application scenarios and draws attention to a challenging but practical research problem: low-resource NER, where
the model is built to quickly identify new entities in a completely unseen target domain with only a few supporting samples in the new domain.
Overall, low-resource NER~\cite{label-agnostic,nearest-neighbor-crf,example-ner} mainly faces two issues as shown in Figure \ref{fig:example}: (1) \textbf{Class Transfer}.
Entity categories can be different across rich-resource settings and low-resource settings.
For example, source news domain contains the entity categories with ``person'' ,``location'' , etc, while target movie domain adds new categories with ``rating\_average'' and ``actor''. In this situation, the current mainstream method such as LC, MRC and BARTNER have to refactor a new model and train it from scratch, which is expensive, and unrealistic for real-world settings.
(2) \textbf{Domain Transfer}. 
Compared with rich-resource settings, the low-resource setting may have a different textual domain. Intuitively, the sentence in news domain and atis domain contain the  different grammar style and allegorical theme, which is not trivial to transfer the model fully trained in the source domain to target domain with few examples.

To address the issue of class transfer, we first reformulate the NER task from sequence labeling to a generative framework with a unified learnable verbalizer to realize \textbf{class transfer}.
Considering different categories involve varying numbers of words as their descriptions,  vanilla mainstream method that assign single classifier for entity may lose important label semantic information. 
Thus, we propose to  construct a unified learnable verbalizer based on generative framework.
Different from BartNER~\cite{unifiedNER}) that has extra MLP layer in Encoder and classifier in Decoder,  
our method only contain the original architecture of pre-trained generative model by constructing constructing a unified learnable verbalizer for entity.
Therefore, our approach can directly leverage any new or complicated entity types without modifying the network structure.

Recently, prompt-tuning~\cite{schick2020exploiting,gao2020making,li2021prefix,DBLP:journals/corr/abs-2103-10385} has emerged to become surprisingly effective for the model adaptation of PLMs, especially in the low-resource setting.
However, the prompt-tuning relies on reformulating the paradigm of downstream tasks into new tasks similar to MLM pre-training, which is not efficient for sequence labeling tasks such as NER. 
Inspired by the success of prompt-learning~\cite{DBLP:conf/emnlp/LesterAC21} in domain adaptation, we propose a lightweight tuning paradigm with pluggable guidance for NER (LightNER) to tackle the downsides of \textbf{domain transfer}.
Specifically, we propose to incorporate learnable parameters into the self-attention layer in LMs and regard the parameters as knowledgeable guidance.
In particular, we explore lightweight tuning with the pluggable guidance module to urge it to learn domain transfer ability and condition it at inference time.

In light of the  limits of the existing techniques, we are interested in building a lightweight tuning framework fot low-resource NER with pluggable prompting.
Notably, the modules in {\ours} are extremely coupled and indispensable to each other.
It is precisely because we design a generative model equipped with a decoupling space to solve the issue class transfer that the pluggable guidance module can realize domain knowledge transfer with lightweight tuning.
In a nutshell, LightNER consists of the following contributions:

\begin{itemize}

\item We convert sequence labeling to the generative framework and construct decoupling space without any label-specific layers to solve the issue of class transfer. Therefore, the proposed method does not require to build a new model from scratch to adapt to a target domain with new entity categories.

\item 
We propose to incorporate learnable parameters into the self-attention layers as pluggable guidance,
which can be seamlessly plugged into the pre-trained  generative models to conduct lightweight tuning with cross-domain and cross-task knowledge transfer ability. Therefore, {\ours} doesn't need to maintain an LM for each target domain NER tasks and pay for expensive training services.

\item We conduct extensive experiments on several benchmark datasets, and by tuning only little parameters, {\ours} can achieve comparable results in standard supervised settings and yield promising performance in low-resource settings.
Our results also suggest that {\ours}
has the potential towards cross domain zero-shot generation with pluggable guidance. 
\end{itemize}



\begin{figure*}[t!]
    \centering
            \subfigure[Overview of {\ours}.]{
    \includegraphics[width=0.6\textwidth]{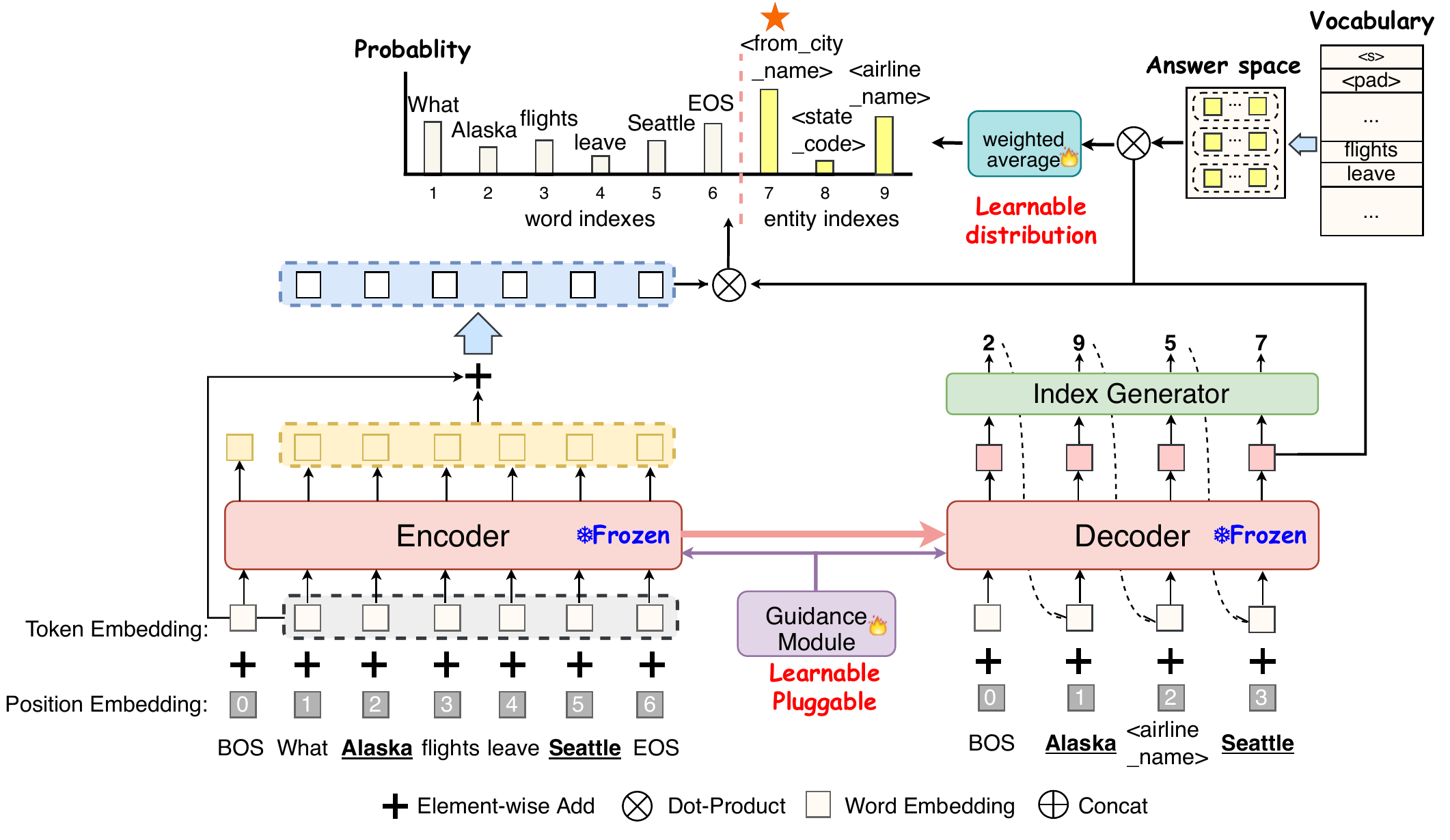}
    \label{fig:arc1}
    }
            \subfigure[Attention with guidance.]{
    \includegraphics[width=0.3\textwidth]{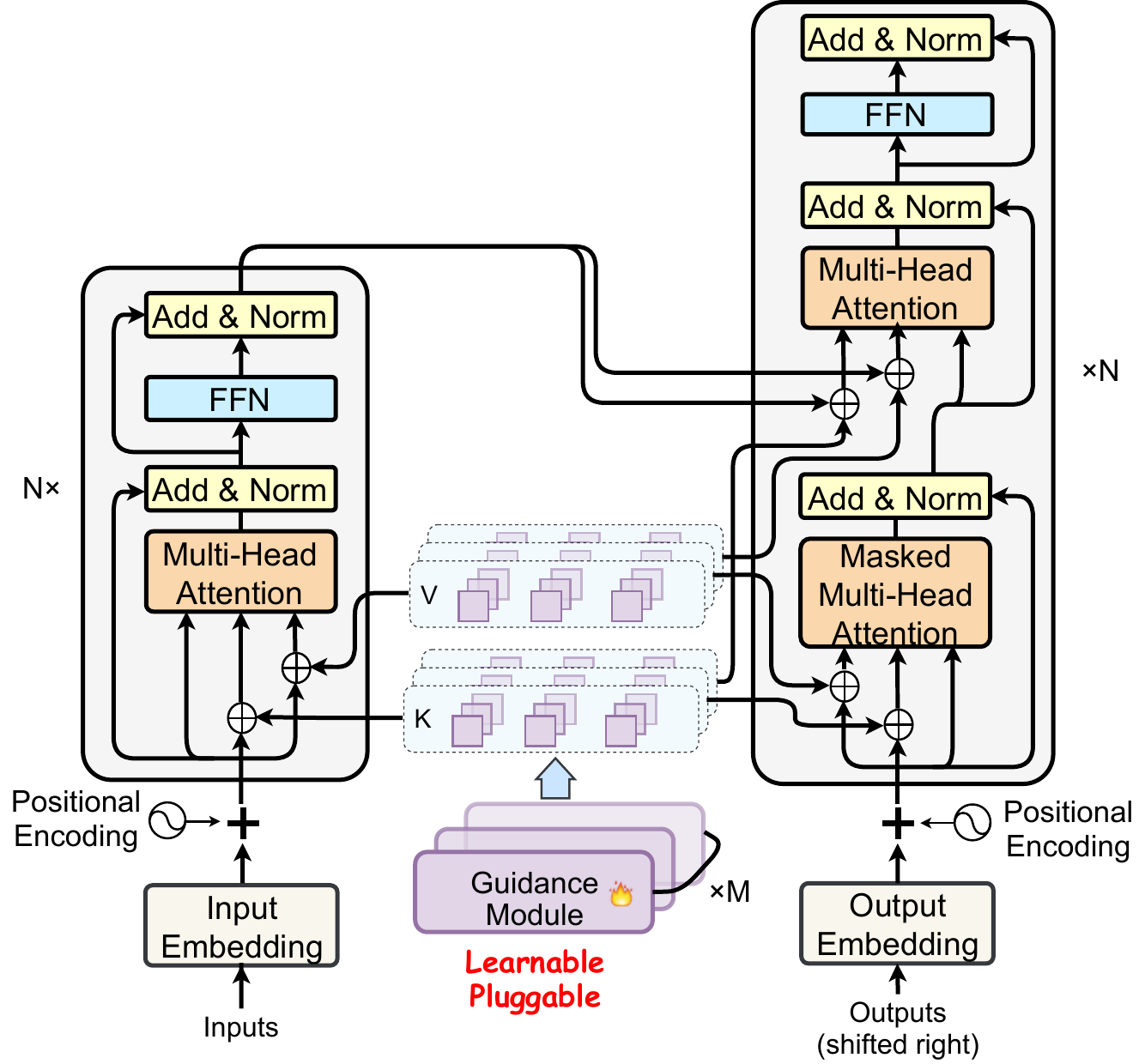}
    \label{fig:arc2}
    }
 
    \caption{Overview of our LightNER framework. 
    \label{fig:model}}
\end{figure*}

%% file: related.tex
\section{Related Work}
\subsection{Named Entity Recognition}
The current dominant methods \cite{lstm-cnn-crf,gcdt,openue,DBLP:conf/naacl/LiuFTCZHG21,noisyner} treat NER as a sequence tagging problem with label-specific classifiers or CRF.
Nevertheless, these works still need to modify the model architecture when facing new entity classes; the inability to solve the challenge of class transfer limits its efficiency and transferability, which is not suitable for low-resource scenarios.
Meanwhile, one crucial research line of low-resource NER  is prototype-based methods, which involve meta-learning and have recently become popular few-shot learning approaches in the NER area. 
Most of the approaches~\cite{DBLP:conf/sac/FritzlerLK19,label-agnostic,nearest-neighbor-crf,example-ner,henderson2020convex,hou2020few,nest1,ding2021few} utilize the nearest-neighbor criterion to assign the entity type, which depends on similar patterns of entity between the source domain and the target domain without fully exploiting the potential of PLMs, behaving unsatisfactorily for cross-domain instances. 

Recently, \citet{DBLP:journals/corr/abs-2106-01760} propose template-based BART for few-shot NER, which enumerates all
$\mathbf{\hat{n}}$-gram possible spans in the sentence and fills them in the hand-crafted templates, classifying each candidate span based on the corresponding template scores.
Different from their approach, our framework does not need template engineering and is more friendly with computation complexity.

\subsection{Prompt Learning for PLMs}

Since the emergence of  GPT-3~\cite{gpt3}, prompt-tuning has received considerable attention. 
A series of research work~\cite{schick2020exploiting, schick2020automatically,shin2020eliciting,han2021ptr,PADA,DBLP:conf/emnlp/PothPRG21,PADA} have emerged, which implies that prompt-tuning can effectively stimulate knowledge from PLMs compared with standard fine-tuning, thus,  inducing better performances on few-shot and cross-domain tasks. 
However, prompt learning mainly focuses on reformulating the downstream tasks' paradigm into completing a cloze task to bridge the gap between pre-training and fine-tuning, lacking an efficient method for NER and other sequence labeling tasks.
Different from recent work of prompt tuning for NER~\cite{plugtagger,template}, we mainly focus on the issues of domain transfer and class transfer for low-resource NER.

\subsection{Lightweight Learning for PLMs}
Lightweight fine-tuning is performed to leverage the ability of PLMs with small trainable parameters. On the one hand, several studies consider removing or masking redundant parameters from PLMs~\cite{FrankleC19,Sanh0R20}. 
On the other hand, 
some researchers~\cite{GuoRK20,ZhangSZGM20} argue that extra trainable modules should be inserted into PLMs. As a typical approach, adapter-tuning~\cite{HoulsbyGJMLGAG19} inserts task-specific layers (adapters) between each layer of PLMs;
prefix-tuning~\cite{li2021prefix} prepends a sequence of continuous task-speciﬁc vectors to the inputs.
However, adapter-tuning adds additional layers into the activation module of LMs, while this modification of the architecture is inconvenient to redeploy the model when switching to a new domain with unseen entity types. Meanwhile, the different label sets among domains make it impossible to transfer prefix-tuning to sequence labeling such as NER.

Apart from Adapter~\cite{HoulsbyGJMLGAG19} and prefix-tuning~\cite{li2021prefix} aiming to conduct efficient finetuning, our approach focus on achieving efficient knowledge transfer through a pluggable paradigm.


%% file: method.tex
\section{Preliminaries}

\subsection{Low-resource NER}
\label{sec:fewshot}
Given a rich-resource NER dataset $\mathbb{H}= \{(\mathbf{X}_1^H,\mathbf{Y}_1^H), ..., (\mathbf{X}_R^H,\mathbf{Y}_R^H)\}$, where the input is a text sequence of length $n$, $\mathbf{X}^H=\{x_1^H,\dots,x_n^H\}$, we use $\mathbf{Y}^H=\{y_1^H,\dots,y_n^H\}$ to denote  corresponding labeling sequence of length $n$, and adopt $\mathcal{C}^H$ to represent the label set of the rich-resource dataset ($\forall y_i^H, y_j^H \in \mathcal{C}^H$). 
Traditional NER methods are trained in the standard supervised learning settings, which usually require many pairwise examples, \ie $R$ is large.
However, only a few labeled examples are available for each entity category in real-world applications due to the intensive annotation cost. 
This issue yields a challenging task of {\it low-resource} NER, in which given a low-resource NER dataset,  $\mathbb{L} = \{(\mathbf{X}_1^L,\mathbf{Y}_1^L), ..., (\mathbf{X}_r^L,\mathbf{Y}_r^L)\}$, the number of labeled data in low-resource NER dataset is quite limited (i.e., $r \ll R$) compared with the rich-resource NER dataset.
Regarding the issues of low resource and cross domain,  the target entity categories $\mathcal{C}^L$ ($\forall l_i^L,l_j^L \in \mathcal{C}^L$) may be different from $\mathcal{C}^H$, which is challenging for model optimization.

\subsection{Label-specific Classifier for NER}
\label{sec:sl}
Traditional sequence labeling methods usually assign a label-specific classifier over the input sequence, which identifies named entities using BIO tags.
A label-specific  classifier with parameter $\thetav = \{\mathbf{W}_{\mathcal{C}}, \mathbf{b}_{\mathcal{C}}\} $  followed by a softmax layer is used to project the representation $\mathbf{h}$ into the label space.
Formally, given $x_{1:n}$, the label-specific  classifier method calculates:
\begin{equation}
\small
\begin{split}
       \mathbf{h}_{1:n} &= \textsc{Encoder}(x_{1:n}), \\
        q(\yv | \xv) &= \textsc{softmax}(\mathbf{h}_i \mathbf{W}_{\mathcal{C}} + \mathbf{b}_{\mathcal{C}}) \ (i \in [1,...,n]),
\end{split}
\label{eq:bert_output}
\end{equation}
where $\mathbf{W}_{\mathcal{C}} \in \mathbb{R}^{d_ \times m}$, $\mathbf{b}_{\mathcal{C}} \in \mathbb{R}^{m}$ are trainable parameters and $m$ is the numbers of entity categories. 
We adopt BERT \cite{DBLP:conf/naacl/DevlinCLT19} and BART \cite{DBLP:conf/acl/LewisLGGMLSZ20} as our \textsc{Encoder} to encoder the representation of text sequence, together with label-specific classifier layer, denoted as  \textbf{LC-BERT} and \textbf{LC-BART} respectively.

\section{Methodology}

\subsection{Task Formulation}
Low-resource NER usually involves the class transfer, where new entity categories exist in target domains; however, the traditional sequence labeling method needs a label-specific output layer based on PLMs, hurting its generalization.
Therefore, we reformulate the NER as a generative framework to maintain the consistency of architecture and enable the model to handle different entity types. 
For a given sentence $X$, we tokenize it into a sequence of tokens $X=\{x_1, x_2, ...x_n\}$.
The NER task aims to provide the start and end index of an entity span, along with the entity type, represented by \emph{e}, \emph{t} in our framework, respectively. 
\emph{e} is the index of tokens and $\emph{t}\in \text{\{``{\it person}'', ``{\it organization}'', ..., \}}$ is the set of entity types. 
Superscript \textsuperscript{start} and \textsuperscript{end} denote the start and end index of the corresponding entity token in the sequence.
For the generative framework, the target sequence $Y$ consists of multiple base prediction $p_i = \{e^{start}_i, e^{end}_i, t_i\}$ and $Y = \{p_1, p_2, ...., p_l\}$, where $l$ denotes num of entities in $X$.
We take a sequence of tokens $X$ as input and hope to generate the target sequence $Y$ as defined above. 
The input and output sequence starts and ends with special tokens
``$<$s$>$'' and ``$<$/s$>$''. 
They should also be generated in $Y$, but we ignore them in equations for simplicity. 
Given a sequence of tokens $X$, the conditional probability is calculated as:
\begin{equation}
\small
    P(Y|X) = \prod_{t=1}^{3l} p(y_t|X, y_0, y_1,...,y_{t-1}).
\end{equation}

\subsection{Generative Framework}
To conduct \textbf{class transfer}, we adopt the seq2seq architecture with the pointer network to model the conditional probability $P(Y|X)$, where the conduction of pointer network is inspired by the ~\citet{DBLP:conf/acl/SeeLM17,unifiedNER}.
Our generative module is shown in Figure~\ref{fig:model}, consisting of two components:
\subsubsection{Encoder}
The encoder is to encode $X$ into the hidden representation space as a vector $H_{en}$. 
\begin{equation}
\small
\begin{aligned}
    H_{en} = \mbox{Encoder}(X)
\end{aligned}
\end{equation}
where $H_{en} \in \mathcal{R}^{n \times d}$ and $d$ is the hidden state dimension.

\subsubsection{Decoder}
The decoder part takes the encoder outputs $H_{en}$ and previous decoder outputs $y_1, y_2,...,y_{t-1}$ as inputs to decode $y_t$. 
${y_i}_{i=1}^{t-1}$ indicates the token indexes; an index-to-token converter is applied for conversion.

\begin{equation}
\small
\begin{split}
\tilde{y}_i= \left\{ 
\begin{array}{ll}
X_{y_i}, & \mbox{if\: $y_i$\: is\: a\: pointer\: index}\\
C_{y_i-n}, & \mbox{if\: $y_i$\: is\: a\: class\: index}
\end{array}
\right.
\end{split}
\end{equation}
where $C=[c_1, c_2,....c_m]$ is the set of entity categories (such as ``Person'', ``Organazation'', etc.), which are answer words corresponding to the entity category\footnote{The index of entity categories always starts after the pointer indexes of the given sequence, at $n + 1$.}.
After this, we then get the last hidden state for $y_t$ with the converted previous decoder outputs $[\tilde{{y}_i}_{i=1}^{t-1}]$.

\begin{equation} 
\small
\begin{aligned}
h_t = \mbox{Decoder}(H_{en};[\tilde{{y}_i}_{i=1}^{t-1}])
\end{aligned}
\end{equation}
where $h_t \in \mathcal{R}^{d}$; moreover, the probability distribution $p_t$ of token $y_t$ can be computed as follows:

\begin{eqnarray}
\small
\begin{aligned}
    E_{seq} &= \mbox{WordEmbed}(X), \\
    \tilde{H}_{en} &= \alpha \cdot H_{en} + (1-\alpha)\cdot E_{seq}, \\
    p_{seq} &= \tilde{H}_{en} \otimes h_t,  \\
    p_t &= \mbox{Softmax}([p_{seq};p_{tag}]),
\end{aligned}
\end{eqnarray}
where $E_{seq}, \tilde{H}_{en}  \in \mathcal{R}^{n \times d}$; $\alpha \in{\mathcal{R}}$ is a hyper-parameter;
$p_{seq}$ and $p_{tag}$ refer to the predicted logits on  index of entity span and entity categories respectively;
$p_t \in \mathcal{R}^{(n+m)} $ is the predicted probability distribution of $y_t$ on all candidate indexes;
$[\, \cdot \, ; \, \cdot \, ]$ denotes concatenation in the first dimension.
In particular, the details of $p_{tag}$ are in the following subsection.

\subsection{Unified Learnable Verbalizer}

As for the prediction of entity categories in NER, it is challenging to manually find appropriate tokens in the vocabulary to distinguish different entity types. 
Besides, some entity type may be complicated or very long in the specific target domain, such as $return\_date.month\_name$ in ATIS \cite{atis} and $restaurant\_name$ in MIT Restaurant \cite{mit-dataset}. 

To address the above issues in class transfer, we construct a unified learnable verbalizer containing multiple label words related to each entity class and leverage the weighted average approach for the utilization of the decoupling space $\mathcal{V}$.
Concretely, we define a mapping $\mathcal{M}$ from the label space of entity categories $\mathcal{C}$ to the unified learnable verbalizer $\mathcal{V}$,
i.e., $\mathcal{M}\colon \mathcal{C} \mapsto \mathcal{V}$. 
We utilize $\mathcal{V}_c$ to represent the subset of $\mathcal{V}$ that is mapped by a specific entity type $c$,  $\mathcal{V} = \cup_{c\in\mathcal{C}} \mathcal{V}_c$.
Take the above $c=\text{``return\_date.month\_name''}$ as example, we set $\mathcal{V}_{c}=\{\text{``return'',``date'',``month'',``name''}\}$ according to decomposition of $c$. 
Since the direct average function may be biased, we adopt learnable weights $\beta$ to average the logits of label words in answer space as the prediction logit:
\begin{equation}
\small
\begin{aligned}
    E_{tag} &= \mbox{WordEmbed}(\mathcal{M}(\mathcal{C})), \\
    p_{tag} &= Concat[\sum_{v \in \mathcal{V}_{c}} {\beta}_v{^c} * E_{tag}^c \otimes h_t]  
\end{aligned}
\end{equation}
where ${\beta}_v{^c}$ denotes the weight of entity type $c$; $\sum_{v \in \mathcal{V}_{c}} {\beta}_v{^c}=1$; $p_{tag} \in \mathcal{R}^{m}$.
Through the construction of the unified learnable verbalizer, {\ours} can perceive semantic knowledge in entity categories without modifying the PLM.

\subsection{Pluggable Guidance Module}

\subsubsection{Parameterized Setting}
Specifically, {\ours} adds two sets of trainable embedding matrices
$\{ \phi^{1}, \phi^{2}, .., \phi^{N} \}$ for the encoder and decoder, respectively, and sets the number of transformer layers as $N$, where
$\phi_\theta \in \mathbb{R}^{{2\times|P|}\times{d}} $ (parameterized by $\theta$), ${|P|}$ is the length of the prompt, $d$ represents the $dim(h_t$), and $2$ indicates that $\phi$ is designed for the key and value.
In our method, the LM  parameters are fixed, and the prompt parameters $\theta$ and the learnable distribution of $\beta$ are the only trainable parameters. 
\subsubsection{Pluggable Guidance Layer}
{\ours} inherits the architecture of the transformer~\cite{vaswani2017attention}, which is a stack of identical building blocks wrapped up with a feedforward network, residual connection, and layer normalization.
As a specific component, we introduce the pluggable guidance layer over the original query/key/value layer to achieve flexible and effective knowledge transfer.
Given an input token sequence $X = \{x_1, x_2, ..., x_n\}$, following the above formulation, we can incorporate the representation of the guidance module into $x$ with the calculation of self-attention.
In each layer $l$, the input sequence representation $\boldsymbol{X}^l\in\mathbb{R}^{n d}$ is first projected into the query/key/value vector:
\begin{equation}
\small
\boldsymbol{Q}^l=\boldsymbol{X}^l\boldsymbol{W}^Q, \boldsymbol{K}^l=\boldsymbol{X}^l\boldsymbol{W}^K, \boldsymbol{V}^l=\boldsymbol{X}^l\boldsymbol{W}^V,
\end{equation}
where $\boldsymbol{W}_l^Q$, $\boldsymbol{W}_l^K$, $\boldsymbol{W}_l^V\in\mathbb{R}^{d \times d}$.
Then, we can redefine the attention operation as:
\begin{equation}
\small
Attention^{l}={softmax}(\frac{\boldsymbol{Q}^l[{\boldsymbol{\phi}_k^l;\boldsymbol{K}^l}]^T}{\sqrt{d}})
[\boldsymbol{\phi}_v^l;\boldsymbol{V}^l].
\end{equation}
Based on these representations of inputs and pluggable guidance module, we aggregate them and compute the attention scores to guide the final self-attention flow.
Consequently, the guidance module can re-modulate the distribution of attention.

%% file: experiments.tex
\begin{table}[!htbp]
\centering
\footnotesize
\scalebox{0.72}{
\begin{tabular}{l|c|c|c}
\toprule
    {\bf Traditional Models} & {\bf P} & {\bf R} & {\bf F} \\
\midrule
    \citet{yang-etal-2018-design} & - & - & 90.77  \\
    \citet{lstm-cnn-crf} & - & - & 91.21  \\
    \citet{luke}  & - & - & \textbf{94.30}  \\
     \citet{refine} & - & - & 92.02 \\
     \citet{DBLP:conf/acl/LiFMHWL20} $\dagger$ 
     & 92.47 & 93.27  & 92.87  \\
     \citet{DBLP:conf/acl/YuBP20} $\ddagger$ 
     & \textbf{92.85} & 92.15 & 92.50 \\
     LC-BERT 
     & 91.93 & 91.54 & 91.73  \\
     LC-BART 
     & 89.60 & 91.63 & 90.60  \\
\midrule
         {\bf Few-shot Friendly Models} & {\bf P} & {\bf R} & {\bf F} \\
\midrule
     \citet{label-agnostic} 
     & - & - & 89.94 \\
     Template~\cite{DBLP:journals/corr/abs-2106-01760} 
     & 90.51 & 93.34 & 91.90 \\
     \textbf{{\ours}} &92.39 & \textbf{93.48} & 92.93 \\
\bottomrule
\end{tabular} 
}
\caption{Model performance on the CoNLL-2003 dataset \label{table:mp_conll}.
``$\dagger$'' indicates that we rerun their code with BERT-LARGE~\cite{DBLP:conf/naacl/DevlinCLT19}. ``$\ddagger$'' indicates our reproduction with only the sentence-level context. Although  LUKE~\cite{} is pre-trained with a large entity-annotated
corpus (Wikipedia), {\ours} is  highly competitive in rich resource settings even though it is designed for low-resource NER.
}
\end{table}

\section{Experiments}
We conduct extensive experiments in standard and low-resource settings. 
We use CoNLL-2003 \cite{conll03} as the rich-resource domain.
Following the settings in \citet{example-ner} and \citet{huang2020fewshot}, we use the Massachusetts Institute of Technology (MIT) Restaurant Review \cite{mit-dataset}, MIT Movie Review \cite{mit-dataset}, and Airline Travel Information Systems (ATIS) \cite{atis} datasets as the cross-domain low-resource datasets\footnote{We do not conduct experiemnts on Few-NERD \cite{ding2021few} since our setting follows \cite{example-ner} which is different from the N-way K-shot settting.}.
Our experiments are evaluated in an exact match scenario, data analysis, and implementation details are presented in the Section Appendix~\ref{sup:data_analysis} and ~\ref{sup:details}. We also provide the supplementary experimental result for the in-domain low-resource setting as shown in Section Appendix~\ref{sup:indomain}.

\subsection{Standard Supervised NER Setting}
We adopt the CoNLL-2003 dataset to conduct experiments in the standard supervised settings.
A comparison of the results of {\ours} and the SOTA methods are listed in Table~\ref{table:mp_conll}. 
Mainly, LC-BERT and LC-BART provide a strong baseline.
We identify that even though {\ours} is designed for the low-resource NER, it is highly competitive with the best-reported score in the rich-resource setting as well,
indicating the effectiveness of our decoding strategy and guidance module. 

\begin{table*}[t!]
\centering
\small
\scalebox{0.71}{
\begin{tabular}{c|l|cccccc|cccccc|ccc}
\toprule

{\multirow{2}{*}{Source}} 
& {\multirow{2}{*}{Methods}} 
& \multicolumn{6}{c|}{\textit{MIT Movie}}
& \multicolumn{6}{c|}{\textit{MIT Restaurant}}
& \multicolumn{3}{c}{\textit{ATIS}}\\
\cmidrule{3-17}
 &  &10 &20 &50 &100 &200 &500 &10 &20 &50 &100 &200 &500  &10 &20 &50\\

\midrule

    \multirow{4}{*}{None} 
    & LC-BERT~\  & 25.2 & 42.2 & 49.6 & 50.7 & 59.3 & 74.4 
    & 21.8 & 39.4 & 52.7 & 53.5 & 57.4 & 61.3
    & 44.1 & 76.7 & 90.7\\
    & LC-BART~\  & 10.2 & 27.5 & 44.2 & 47.5 & 54.2 & 64.1 
    & 6.3 & 8.5 & 51.3 & 52.2 & 56.3 & 60.2
    & 42.0 & 72.7 & 87.5\\
    & Template & 37.3 & 48.5 & 52.2 & 56.3 & 62.0 & 74.9 
    & 46.0 & 57.1 & 58.7 & 60.1 & 62.8 & 65.0
     & 71.7 & 79.4 & 92.6\\
    & BERT-MRC$\dagger$  & 18.7 & 48.3 & 55.5 & 62.5 & 80.2 & 82.1
    & 12.3 & 37.1 & 53.5 & 63.9 & 65.5 & 70.4
    & 35.3  & 63.2  & 90.2 \\
\cmidrule{2-17}

    & \textbf{\ours}  & \textbf{41.7} & \textbf{57.8} & \textbf{73.1} & \textbf{78.0} & \textbf{80.6} & \textbf{84.8} 
    & \textbf{48.5} & \textbf{58.0} & \textbf{62.0} & \textbf{70.8} & \textbf{75.5} & \textbf{80.2}
    & \textbf{76.3} & \textbf{85.3} & \textbf{92.8} \\

\midrule
    \multirow{6}{*}{CoNLL03} 
    & Neigh.Tag. &\ 0.9 &\ 1.4 &\ 1.7 &\ 2.4 &\ 3.0 &\ 4.8 
    &\ 4.1 &\ 3.6 &\ 4.0 &\ 4.6 &\ 5.5 &\ 8.1
    &\ 2.4 &\ 3.4 &\ 5.1\\
    & Example. & 29.2 & 29.6 & 30.4 & 30.2 & 30.0 & 29.6 
    & 25.2 & 26.1 & 26.8 & 26.2 & 25.7 & 25.1
    & 22.9 & 16.5 & 22.2\\
    & MP-NSP  & 36.4 & 36.8 & 38.0 & 38.2 & 35.4 & 38.3 
    & 46.1 & 48.2 & 49.6 & 49.6 & 50.0 & 50.1
    & 71.2 & 74.8 & 76.0\\
    & LC-BERT & 28.3 & 45.2 & 50.0 & 52.4 & 60.7 & 76.8 
    & 27.2 & 40.9 & 56.3 & 57.4 & 58.6 & 75.3
    & 53.9 & 78.5 & 92.2\\
    & LC-BART  &13.6&30.4&47,8&49.1&55.8&66.9
    &8.8&11.1&42.7&45.3&47.8&58.2
    &51.3&74.4&89.9\\
    & Template & 42.4 & 54.2 & 59.6 & 65.3 & 69.6 & 80.3
    & 53.1 & 60.3 & 64.1 & 67.3 & 72.2 & 75.7
    & 77.3 & 88.9 & 93.5\\
    & BERT-MRC$\dagger$  & 20.2  & 50.8 & 56.3 & 62.9 & 81.5 & 82.3
    & 15.8 & 39.5  & 54.8 & 65.8 & 68.8  & 73.5
    &40.5  &66.7  &91.8 \\
\cmidrule{2-17}

    & \textbf{\ours} 
    & \textbf{62.9}  & \textbf{75.6}  & \textbf{78.8} & \textbf{82.2} & \textbf{84.5} & \textbf{85.7} 
    & \textbf{58.1} & \textbf{67.4} & \textbf{69.5}& \textbf{73.7} & \textbf{78.4} & \textbf{81.1}
    & \textbf{86.9} & \textbf{89.4} & \textbf{93.9}\\

\bottomrule
  
\end{tabular}
}
\caption{\label{tab:cross}
Model performance (F1 score) in the cross-domain low-resource setting. ``$\dagger$'' indicates that we rerun their public code in this setting.
All of our experiments and baselines adopt large version of LMs.}
\end{table*}

\subsection{Cross-Domain Low-resource NER Setting} 
In this section, we evaluate the model performance in the scenarios in which the target entity categories and textual style are specifically different from the source domain, and only limited labeled data are available for training.
Precisely,  we follow the setting in \citet{DBLP:journals/corr/abs-2106-01760} to  sample a specific number of samples per entity category randomly as the training data in the target domain to simulate the cross-domain low-resource data scenarios.
Table \ref{tab:cross} lists the results of training models on the CoNLL-2003 dataset as a generic domain and its evaluations on other target domains.
The results of {\ours} are based on running the experiments five times on random samples and calculating the average of their scores.


\paragraph{Competitive Baselines}
We consider seven competitive approaches in our experiments.
The {\it prototype-based methods}\footnote{Note that even if the prototype-based methods is training-free in the target domain, they are by no means equivalent to zero-shot setting, since prototype-based methods require labeled data in target domain as supporting examples.} primarily include the following:
$(\RN{1})$
{\it Neigh.Tag.}~\cite{label-agnostic};
$(\RN{2})$
{\it Example-based NER}~\cite{example-ner};
$(\RN{3})$
{\it Multi-prototype + NSP } (referred to as {\it MP-NSP })~is a SOTA prototype-based method reported in~\cite{huang2020fewshot}, utilizing noisy supervised pretraining. 
The {\it label-specific classifier} mainly include the following: 
$(\RN{4})$ {\it LC-BERT} and $(\RN{5})$
{\it LC-BART} is the adoption of the 
label-specific classifiers on top of corresponding PLMs.
Besides, $(\RN{6})$
{\it Template-based BART}~\cite{DBLP:journals/corr/abs-2106-01760} recently propose a template-based method for few-shot NER and $(\RN{7})$
{\it BERT-MRC}~\cite{DBLP:conf/acl/YuBP20} propose to formulate NER as a machine reading comprehension (MRC) task, which is a strong SOTA model for NER. A summary comparison with baselines is shown in Section 4 of Appendix.


\paragraph{Train from Scratch on Target Domain} 
We first consider direct training on the target domain from scratch without any available source domain data.
However, prototype-based methods cannot be used in this setting. 
When compared to the LC-BART, LC-BERT,  template-based BART and BERT-MRC,  the results of our approach is consistently more persistent, indicating {\ours} can better exploit few-shot data.
Particularly, {\ours} achieve an F1-score of 57.8\% in 20-shot setting on MIT Movie, which is higher than the results of LC-BERT and template-based BART in 50-shot setting.

\paragraph{Transfer Knowledge from a General Domain to Specific Domains}
We observe that the performance of prototype-based methods remains approximately the same as the number of labeled data increases,
while {\ours} continues to improve when the number of target-domain data increases.
Table~\ref{tab:cross} shows that on all three target-domain datasets, {\ours} significantly outperforms the other three types of baselines in the case of both 10 and 500 instances per entity type,
From the perspective of quantifying the \textbf{knowledge transferred}, when the number of instances is 10, the performance of our model increase the F1-scores to 21.2\%, 9.6\%, and 10.6\% on the MIT movie, MIT restaurant, and ATIS datasets, respectively, which are better than the results of knowledge transferred by {\it LC-BERT}. 
This demonstrates that our model is more successful in transferring the knowledge learned from the source domain. 

\begin{table}[!htb]
\centering
\footnotesize
\scalebox{0.72}{
\begin{tabular}{c|l|ccc}
\toprule
{\multirow{2}{*}{Source}} & 
{\multirow{2}{*}{Methods}} 
& \multicolumn{3}{c}{\textit{MIT Restaurant}}
\\
\cmidrule{3-5}
 &  &10 &20 &50 \\

\midrule
    \multirow{5}{*}{None} 
    & Ours [BART] & 48.5 & 58.0 & 62.0
    \\
     & \tableindent- pluggable module    
     & \textbf{50.3} & \textbf{59.4} & \textbf{63.5}
    \\
     & \tableindent- unified learnable verbalizer
     & 45.5 & 55.5 & 59.8
    \\
\cmidrule{2-5}
    & Ours (Full-params Tuning)   & 49.5 & 59.0 & 62.8
    \\
    & LC-BERT~\   & 21.8 & 39.4 & 52.7 
    \\
    &  LC-BERT+[P-tuning] ~\  & 24.9 & 41.2 & 53.5
   \\
    & LC-BERT+[Adapter]~\ & 11.5 & 14.3 & 21.2
   \\
    & Ours+[Adapter]~\ & 43.3 & 52.3 & 58.5
   \\

\midrule
    \multirow{6}{*}{CoNLL03} 
    & Ours [BART]   & \textbf{58.1} & \textbf{67.4} & \textbf{69.5}
    \\
    & \tableindent - pluggable module  
    & 54.5 & 64.2 & 67.8 \\
    & \tableindent - unified learnable verbalizer  
    & 48.7 & 58.8 & 62.5
    \\
\cmidrule{2-5}
    & Ours (Full-params Tuning)  & 53.7 & 63.5 & 66.9 \\
    & LC-BERT~\ & 27.2 & 40.9 & 56.3 \\
    & LC-BERT+[P-tuning] ~\   & 30.3 & 46.8 & 58.2
   \\
    & LC-BERT+[Adapter]~\   & 13.0 & 16.2 & 21.8
   \\
    & Ours+[Adapter]~~\   & 46.8 & 58.2 & 62.5
    \\
   

\midrule
\end{tabular}
}
\caption{\label{tab:ablation}
Performance of Ablation and Variants Study.}
\end{table}

\subsection{Ablation and Comparison}
As shown in the above experiments that our {\ours} possess the outstanding ability of knowledge transfer in the low-resource setting, we demonstrate that the pluggable guidance module contributes to the cross-domain improvement. 
To this end, we ablate the pluggable module and unified learnable verbalizer to validate the effectiveness.
\textit{- pluggable module} indicates the entire parameter (100\%) tuning without our proposed pluggable module. 
\textit{- unified learnable verbalizer} donates our model only randomly assigns one token in the vocabulary to represent the type.
From Table \ref{tab:ablation}, we notice that only \textit{- pluggable module} in the vanilla few-shot setting performs a little better than {\ours}, but \textbf{decreases significantly in the cross-domain few-shot setting}. 
However, \textit{- unified learnable verbalizer} drop both in the two settings.
It further demonstrates that the design of the pluggable module is parameter-efficient and \textbf{beneficial for knowledge transfer}, while unified learnable verbalizer can \textbf{handle class transfer}, which is also essential for low-resource NER.

We further compare {\ours} with several
variants of our method: $(\RN{1})${\it Ours (Full-params Tuning)}; $(\RN{2})${\it LC-BERT+[P-tuning]}: we set the length of continuous template words to be 10 for \textit{ P-tuning}; $(\RN{3})${\it LC-BERT+[Adapter]};  $(\RN{4})${\it Ours+[Adapter]}; 
 \textbf{Firstly}, compared with {\ours},
training all the parameters of our model merely improve a little in in vanilla few-shot setting, but drops significantly in cross-domain few-shot settings, which reveals that our pluggable module with LMs fixed is the vital for transferring knowledge across domains.
\textbf{Secondly}, we observe that LC-BERT equipped with P-tuning achieves a few improvements both in vanilla few-shot and cross-domain few-shot settings.
While Adapter makes performance drop significantly because LC-BERT cannot handle the class transfer, thus the few tuned parameters yield unsatisfactory performance.
\textbf{Finally},  we replace the pluggable module with the Adapter to validate the effectiveness of our module. The fact that {\it Ours+[Adapter]} performs significantly better than {\it LC-BERT+[Adapter]} demonstrates the superiority of our generative framework.
Besides, {\it Ours+[Adapter]} 
behaves unsatisfactorily in a cross-domain low-resource setting, which reveals its poor ability of knowledge transfer for NER.

\begin{table}[t!]
\centering
\small
\scalebox{0.65}{
\begin{tabular}{l|ccc|ccc}
\toprule

{\multirow{2}{*}{Methods}} 
& \multicolumn{3}{c|}{\textit{NER} $\rightarrow$ \textit{POS}} 
& \multicolumn{3}{c}{\textit{POS} $\rightarrow$ \textit{NER}}
\\
\cmidrule{2-7}
 &10 &20  &Full &10 &20  &Full  
 \\

\midrule
    LC-BERT~\  & 44.3/46.2 & 53.7/54.3 & 91.4/91.7 & 37.9/38.3 
    & 48.4/48.6 & 91.7/91.3 
    \\


    \textbf{\ours} & 45.5/50.6 & 54.4/57.7 & 91.3/93.2
    &  46.5/51.8 &  61.8/65.2 & 92.9/93.5
   \\

\bottomrule
\end{tabular}
}
\caption{\label{tab:cross-task}
Model performance in the cross-task setting. Number before and after ``/'' donate the F1 scores of  training from scratch on target domain and transferring from source task to target task respectively.
}
\end{table}

\begin{table}[t!]
\centering
\small
\scalebox{0.75}{
\begin{tabular}{c|ccc|ccc|ccc}
\toprule
    Target & 
     \multicolumn{3}{c|}{CoNLL} & \multicolumn{3}{c|}{Movie} &
     \multicolumn{3}{c}{Restaurant} \\
\midrule
    Source
    & M & R & Mix 
    & C & R & Mix 
    & C & M & Mix \\
\midrule

     LC-BERT & 0.2 & 0.4 & 0.0
       & 0.5 & 0.3 & 0.0
       & 0.3 & 0.2 & 0.0\\
     Template & 0.1 & 0.2  & 0.0
       & 0.3 & 0.2 & 0.0
       & 0.1 & 0.0 & 0.0\\
     \textbf{LightNER} & \textbf{8.5} & \textbf{8.8} &\textbf{15.8}
      & \textbf{12.6} & \textbf{9.0} &\textbf{18.9}
      & \textbf{11.0} & \textbf{8.5} & \textbf{18.4}\\
\bottomrule
\end{tabular}
}
\caption{Cross-domain zero-shot performancefootnote{In zero-shot setting, the weight of the unified learnable verbalizer is average operation.}. C, M, and R refer to the dataset of CoNLL03, Movie, and  Restaurant, respectively. 
The Mix column refers to the methods of averaging the parameters from the other two source domains (average the prompt for {\ours}).  \label{table:zero-shot}}
\end{table}

\subsection{Detailed Model Analysis}
\label{sec:discussion}

\paragraph{The Transferability Across Task} 

Although our {\ours} is designed for NER, it is easy to generalize to other sequence tagging tasks without any modification network structure.

Thus, 
we try to train on full data of the source task, and then simply \textbf{load the pluggable guidance module} to further train the model on the target task.
As shown in Table~\ref{tab:cross-task}, we find 
LC-BERT has an extensive performance drop of all tasks in a cross-task setting, and we believe this is due to the task-specific classifier head hindering the generalization.
The excellent performance in the cross-task setting proves that {\ours} can adapt to other sequence labeling tasks and incredibly transfer knowledge across tasks.



\paragraph{Cross-Domain Zero-Shot Analysis with Mixed Guidance Parameters}
We leverage one dataset as the source domain and conduct the zero-shot experiments on target domains.
From Table~\ref{table:zero-shot}, 
we observe that our method can achieve F1-scores of approximately 10\% in the cross-domain zero-shot setting, significantly higher than other methods.
we further attempt to investigate the performance of mixing different pluggable guidance module.
Specifically, we directly average the parameters of prompts from two source domains as a mixed prompt for the target domain and insert it into the generative framework to evaluate the target performance.
From Table \ref{table:zero-shot}, we notice that mixed prompt achieves promising improvement, which is close to the addition of the results of the original two sources prompt-based model.
We argue that this finding may also inspire future research directions of prompt-tuning and data augmentation.

\begin{figure}
\hspace{-10pt}
    \centering
     \includegraphics[width=0.35\textwidth]{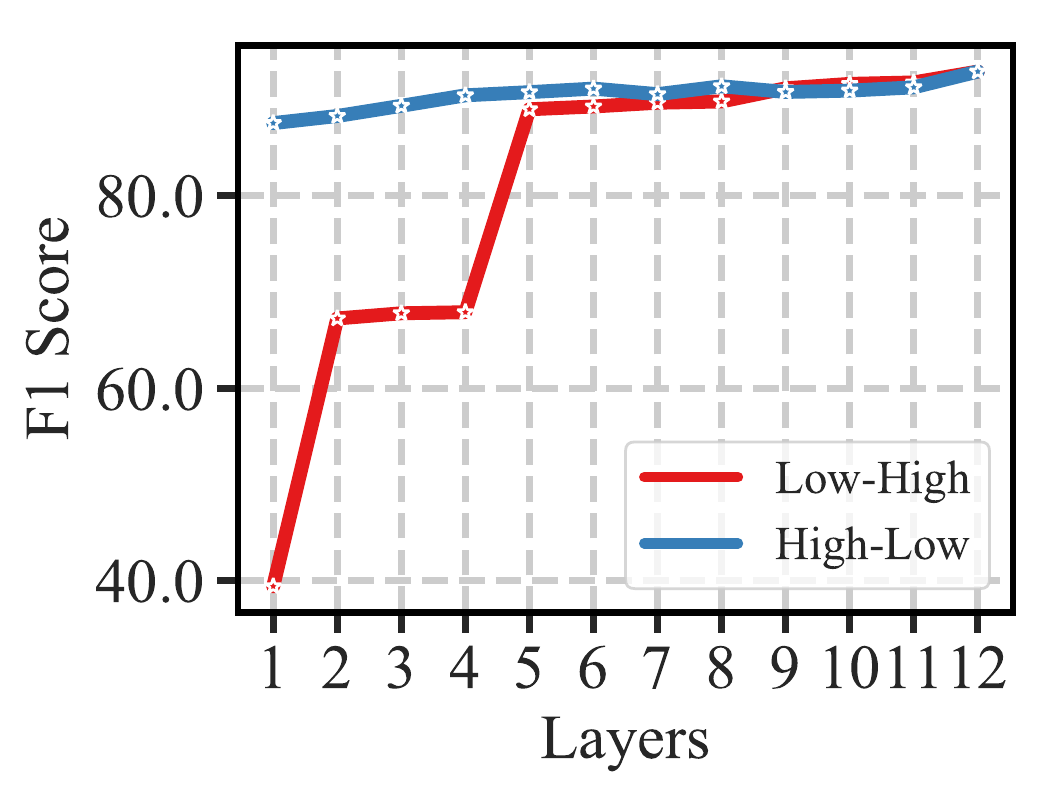}
    \label{fig:layer}
    \caption{\label{fig:len_deep} Performances on CoNLL03 as the layers of guidance module varies.  
    }
\end{figure}

\paragraph{Low-high Layer vs. High-low Layer}

In the aforementioned experiments, we assign the pluggable guidance module to all layers in PLM.
However, it is intuitive to investigate which layer is more sensitive with our approach. 
Intuitively, basic syntactic information may appear earlier in the PLM, while high-level semantic information emerges in higher-level layers \cite{clark2019does}.
We conduct experiments by applying our pluggable module from the lowest to the highest layer and from the highest to the lowest layer separately. 
These two progressive methods are briefly denoted as low-high and high-low, respectively. 
As Figure \ref{fig:len_deep} shows, 
the performance on CoNLL-2003  is close to the original result  obtained after adding full-layer guidance module for tuning. 
This phenomenon also appears in the cross-domain few-shot setting.
(Detailed results refer to Section 3 of  Appendix~\ref{sup:vary}.).
This proves that guidance module applied to higher layers of LMs can better stimulate knowledge from PLMs for downstream tasks more efficiently.


%% file: appendix.tex
 
\section{Detailed Statistics of Datasets}
\label{sup:data_analysis}
We take the standard split of CoNLL03 by following \citet{conll03}, and splits MIT Movie Review, MIT Restaurant Review and ATIS by following \citet{mit-dataset}.
Table~\ref{tab:data_statistic} presents detailed statistics of our datasets.
The standard precision, recall and F1 score are used for model evaluation.

\begin{table}[h!]
    \centering
    \scalebox{0.85}{
    \begin{tabular}{c|c|c|c}
    \hline
         Dataset & \# Train & \# Test & \# Entity \\
    \hline
    CoNLL03 & 12.7k & 3.2k & 4 \\
    MIT Restaurant & 7.6k & 1.5k & 8 \\
    MIT Review & 7.8k & 2k & 12 \\
    ATIS & 4.6k & 850 & 79 \\
    \hline
    \end{tabular}
    }
    \caption{Statistic of datasets.}
    \label{tab:data_statistic}
\end{table}

\begin{table}[!htb]
\centering
\small
\scalebox{0.85}{
\begin{tabular}{l|c|c|c|c|c}
\toprule
    Models & PER & ORG & LOC* & MISC* & Overall \\
\midrule
    LC-BERT & 76.25 & 75.32 & 61.55 & 59.35 & 68.12 \\
    LC-BART & 75.70 & 73.59 & 58.70 & 57.30 & 66.82 \\
    Template & 84.49 & 72.61 & 71.98 & 73.37 & 75.59 \\
    \textbf{\ours} & \textbf{90.96} & \textbf{76.88} & \textbf{81.57} & \textbf{82.08} & \textbf{78.97} \\
\bottomrule
\end{tabular}}
\caption{In-domain low-resource performance on the CoNLL-2003 dataset. * indicates the low-resource entity type. \label{table:conll_few}}
\end{table}

\section{Experimental Details}
\label{sup:details} 
This section details the training procedures and hyperparameters for each of the datasets. 
Considering the instability of the few-shot learning, 
we run each experiment 5 times on the random 
seed [1, 2, 49, 4321, 1234] and report the averaged 
performance. 
We utilize Pytorch to conduct experiments with 1 Nvidia 3090 GPUs. All optimizations are performed with the AdamW optimizer with a linear warmup of learning rate over the first 10\% of gradient updates to a maximum value, then linear decay over the remainder of the training. We set the hyper-parameter $\alpha$ as 0.5. And weight decay on all non-bias parameters is set to 0.01. 
We describe the details of the training hyper-parameters in the following sections.

\subsection{Standard Supervised Setting}
For all models, we fix the batch size as 
16 and search for the learning rates in varied intervals [1e-5, 5e-5].  We train the model for 30 epochs and do evaluation after 20 epoch. We choose
the model performing the best on the validation set 
and evaluate it on the test set.

\subsection{Low-Resource Setting}
We fix the batch size as 
16 and search for the learning rates in varied intervals [3e-5, 5e-5].  We train the model for 30 epochs and do 
evaluation after 20 epoch. We choose
the model performing the best on the validation set 
and evaluate it on the test set.

\subsection{Cross-Task Setting}
We fix the batch size as 
8 and search for the learning rates in varied intervals [2e-5, 5e-5].  We train the model for 30 epochs and do 
evaluation after 25 epoch. We choose
the model performing the best on the validation set 
and evaluate it on the test set.

\section{Supplementary Experimental Results}

\subsection{In-Domain Low-Resource NER Setting} 
\label{sup:indomain} 
Following ~\cite{DBLP:journals/corr/abs-2106-01760}, we construct few-shot learning scenarios on CoNLL-2003 by downsampling, which limits the number of training instances for certain specific categories. 
Particularly, we choose `` LOC'' and ``MISC'' as the low-resource entities and ``PER'' and ``ORG'' as the rich-resource entities. 
The rich and low-resource entity categories have the same textual domain. 
Specifically, we downsample the CoNLL-2003 training set and generate 4,001 training instances, including 2,496 ``PER,'' 3,763 ``ORG,'' 50 ``MISC,'' and 50 ``LOC'' entities.
As shown in Table \ref{table:conll_few}, our method outperforms other methods for both rich- and low-resource entity types. 
This proves that our proposed method has a more substantial performance for in-domain few-shot NER and demonstrates that it can effectively handle the class transfer, which is a challenging aspect in few-shot NER tasks.

\begin{table*}[ht!]
\centering
\small
\scalebox{0.9}{
\begin{tabular}{c|l|ccc|ccc|ccc}
\toprule

{\multirow{2}{*}{Source}} 
& {\multirow{2}{*}{Methods}} 
& \multicolumn{3}{c|}{\textit{MIT Movie}}
& \multicolumn{3}{c|}{\textit{MIT Restaurant}}
& \multicolumn{3}{c}{\textit{ATIS}}\\
\cmidrule{3-11}
 &  &10 &20 &50 &10 &20 &50  &10 &20 &50\\

\midrule

    \multirow{4}{*}{None} 
    & Template & 37.3 & 48.5 & 52.2
    & 46.0 & 57.1 & 58.7 & 71.7 & 79.4 & 92.6\\
    & LightNER(lowest 1 layer) & 16.3 & 20.3 & 30.5
    & 14.6 & 23.4 & 25.4 & 30.6 & 38.3 & 44.2\\
    & LightNER(highest 1 layer)& 29.5 & 38.4 & 45.5
    & 35.4 & 45.3 & 50.5 & 60.1 & 69.8 & 78.7\\
     & LightNER(highest 6 layers)   & 38.5 & 50.5 & 69.8
    & 44.3 & 55.7 & 59.8 & 70.2 & 80.2 & 88.4\\
\cmidrule{2-11}

    & \textbf{LightNER(all 12 layers)}  & \textbf{41.7} & \textbf{57.8} & \textbf{73.1}
    & \textbf{48.5} & \textbf{58.0} & \textbf{62.0}
    & \textbf{76.3} & \textbf{85.3} & \textbf{92.8} \\
\midrule
    \multirow{4}{*}{CoNLL03} 
    & Template & 42.4 & 54.2 & 59.6
    & 53.1 & 60.3 & 64.1 & 77.3 & 88.9 & 93.5\\
    & LightNER(lowest 1 layer) & 24.3 & 30.5 & 35.4
    & 15.6 & 22.4& 27.5 & 37.9 & 44.5 & 48.3\\
    & LightNER(highest 1 layer)& 44.6 & 59.3 & 74.3
    & 39.4 & 45.2 & 51.7 & 59.7 & 68.5 & 79.2\\
     & LightNER(highest 6 layers)   & 55.8 & 69.7 & 75.8 
    & 50.7 & 62.7 & 66.7 & 79.2 & 86.3 & 91.8\\
\cmidrule{2-11}

    & \textbf{LightNER(all 12 layers)} 
    & \textbf{62.9}  & \textbf{75.6}  & \textbf{78.8}
    & \textbf{58.1} & \textbf{67.4} & \textbf{69.5}
    & \textbf{86.9} & \textbf{89.4} & \textbf{93.9}\\
\bottomrule
  
\end{tabular}
}
\caption{\label{tab:cross-layer}
Performances in cross-domain low-resource setting as the prompt layer varies.
}
\end{table*}

\subsection{The Performance in the Low-Resource Setting When the Prompt Layer Varies}
\label{sup:vary}
Intuitively, basic syntactic information may appear earlier in the PLM, while high-level semantic information emerges in higher-level layers. 
Table~\ref{tab:cross-layer} shows that the performance of prompts within highest 1 layer is better than lowest 1 layer overall, and the performance of  highest 6 layers is close to the result of all 12 layers. This proves that prompts applied to higher layers of LMs can better stimulate knowledge from PLMs for downstream tasks more efficiently.

\begin{figure}[ht!]
    \centering
    \includegraphics[width=0.47\textwidth]{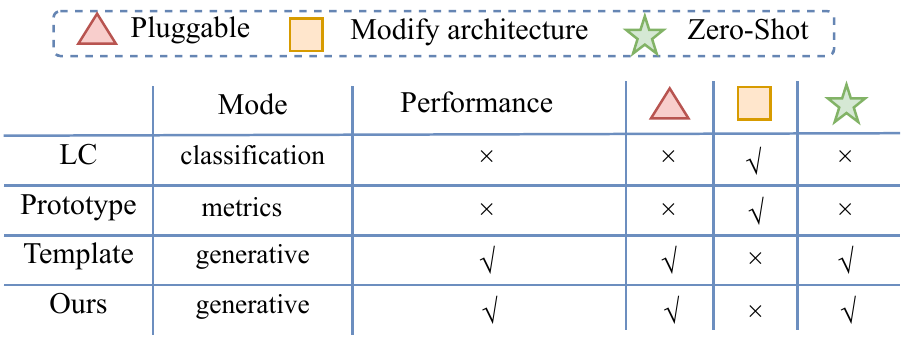}
    \caption{We show the formulations of different NER models and illustrate their corresponding strengths. Zero-shot refers to zero-shot learning ability; LC is short for label-specific classifier (vanilla sequence labeling)}
    \label{fig:comparison}
\end{figure}



\begin{algorithm}[h!]
  \begin{algorithmic}[1]
    \caption{Decoding Algorithm to Convert the Entity Index Sequence into Entity Spans}
    \Require $n$, the number of tokens in $X$; $m$, the number of entity types;  target sequence $Y=[y_1, ..., y_{3l}]$, $l$ is the number of entities; and we have $y_t \in [1, n+m]$
    \Ensure  Entity spans $E=\{(e^{start}_1, e^{end}_1, t_1),...,(e^{start}_i, e^{end}_i, t_i)
    \}$    
    
    \State $E=\{\}, e=[], i=1$
        \While{$i<=3l$}
          \State $y_i = Y[i]$
          \If{$y_i>n$}
          \State  $E.add((e, C_{y_i-n}))$
          \State $e=[]$
          \Else
          \State $e.append(y_i)$
          \EndIf
          \State $i+=1$
        \EndWhile
    \State \Return{$E$}
  \label{alg:decode}
  \end{algorithmic}
\end{algorithm}

\subsection{Impact of Length of Guidance Module}
\begin{figure}
\hspace{-10pt}
    \centering
    \includegraphics[width=0.28\textwidth]{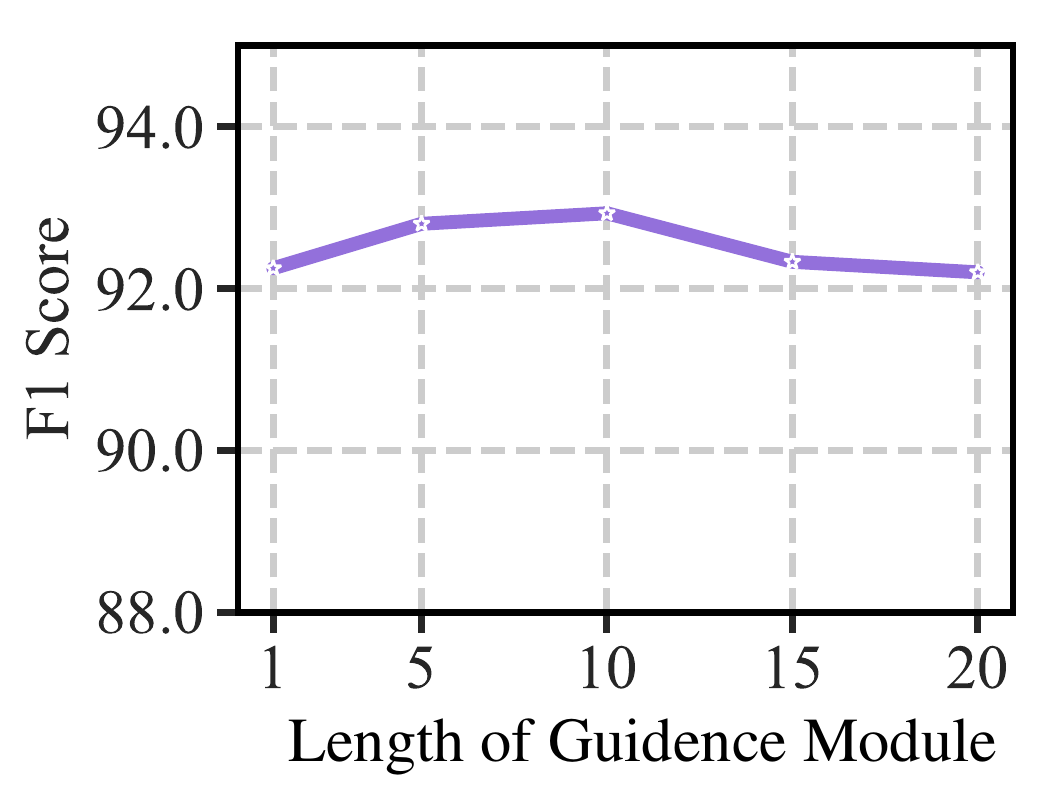}
    \caption{\label{fig:len} Performances on CoNLL03 as the length of guidance module varies.  
    }
\end{figure}

We set the length of prompts as 10 in the above experiment and analyze whether the impact of the length of the prompt. 
From Figure \ref{fig:len}, we notice that a longer prompt implies more trainable parameters but does not guarantee more expressive power. 
It also reveals that our pluggable guidance module is stable; as the length changes, the performance fluctuation does not exceed 1\%.

\section{Comprehensive Comparison}
\label{sup:comparison}
We carry out a comprehensive comparison with related methods as shown in Figure~\ref{fig:comparison}.
For a given sequence, the computational complexity of our LightNER is $O\left(n^{2} d\right)$ and Tamplate-based BART is $O\left(nm\hat{n}\cdot n^{2} d\right)$,where $d$ donates the dimension of the LMs; $\mathbf{n, m, \hat{n}}$ imply the length of input, number of entity classes and  n-grams, respectively. 
Note that our approach does not need to enumerate all possible spans and construct templates, which is efficient than the Template-based method \cite{DBLP:journals/corr/abs-2106-01760}. 
Moreover, we only tune  2.2\%  parameters of the whole model (the tuned params divided by params of the LM), making it memory efficient during training.

\section{The decoding algorithm for converting process}
The decoding algorithm for converting the predicted index sequence to entity spans is shown in Algorithm \ref{alg:decode}.

\section{Sampling strategy in low-resource setting}

\subsection{Cross-Domain}

We simulate the cross-domain low-resource data scenarios by random sampling training instances from a large training set as the
training data in the target domain. We use different numbers of instances for training, randomly sampling a number of instances per entity type (10, 20, 50, 100, 200, 500 instances per entity tag
for MIT Movie and MIT restaurant, and 10, 20, 50 instances per entity tag for ATIS). For different instances per entity tag, we sample five times on the random seed [1, 2, 49, 4321, 1234] and report the averaged performance.

In order to alleviate the problem that an instance usually contains multiple entities, we first sort the entity tags according to the number of instances included. Then we sample instances in the sequence of the sorted order. After once sampling, we will update the status(the remaining number of instances to be sampled) of the entity tags. If an entity tag exceeds the limit after the sampling, discard this sampling.

\subsection{Cross-Task}

Since the CoNLL-2003 dataset contains both entity tag and POS information, we use the same sampling data in cross-task setting. For different instances per entity tag, we sample five times on the random seed [1, 2, 49, 4321, 1234] and report the averaged performance.